\documentclass[runningheads]{llncs}
\usepackage{graphicx}
\usepackage{caption}
\usepackage{subcaption}
\usepackage{xspace}
\usepackage[hidelinks, colorlinks=true]{hyperref}

\usepackage{multirow}
\usepackage{booktabs}
\newcommand{\lidar}{LiDAR\xspace}

\begin{document}

\title{ GAN-based \lidar Intensity Simulation\thanks{Richard Marcus was supported by the Bayerische Forschungsstiftung (Bavarian Research Foundation) AZ-1423-20. This version of the contribution has been
accepted for publication, after peer review but is not the Version of
Record and does not reflect post-acceptance improvements, or any corrections. The
Version of Record is available online at: \url{https://doi.org/10.1007/978-3-031-39059-3_28}. Use of this
Accepted Version is subject to the publisher’s Accepted Manuscript terms of use
\url{https://www.springernature.com/gp/open-research/policies/accepted-manuscript-terms}.}
}
%
%
\author{Richard Marcus\inst{1}\orcidID{0000-0002-6601-6457} \and
Felix Gabel\inst{1}\orcidID{0009-0004-9599-4411} \and
Niklas Knoop\inst{2} \and
Marc Stamminger\inst{1}\orcidID{0000-0001-8699-3442}
} 
\authorrunning{R. Marcus et al.}
%
\institute{Chair of Visual Computing, Friedrich-Alexander-Universität Erlangen-Nürnberg, Germany\\
\email{{richard.marcus, felix.gabel, marc.stamminger}@fau.de}\\
 \and
e:fs TechHub GmbH,  Germany\\
\email{niklas.knoop@efs-techhub.com}}
\maketitle              

\begin{abstract}
Realistic vehicle sensor simulation is an important element in developing autonomous driving. As physics-based implementations of visual sensors like \lidar are complex in practice, data-based approaches promise solutions. Using pairs of camera images and \lidar scans from real test drives, GANs can be trained to translate between them.
For this process, we contribute two additions. First, we exploit the camera images, acquiring segmentation data and dense depth maps as additional input for training.
Second, we test the performance of the \lidar simulation by testing how well an object detection network generalizes between real and synthetic point clouds to enable evaluation without ground truth point clouds.
Combining both, we simulate \lidar point clouds and demonstrate their realism.

\end{abstract}

\section{Introduction}
Autonomous Driving is increasingly getting more attention in the automotive industry and research.
A big part of the success depends on the quality of perception systems.
 \lidar is a very important sensor for perception as it provides highly accurate distance measurements.
 Even though the capabilities of \lidar sensors are increasing rapidly, they are still expensive by themselves and acquiring training data in large quantities is difficult.

Simulation environments provide a possible solution, since generating synthetic data to train perception systems is much cheaper than using real data and also allows the simulation of rare or dangerous situations.
However, the quality of the simulated data is often not realistic enough to transfer the trained perception system to the real world.
For the simulation of LiDAR sensors, ray tracing can be employed.
But for this to generate realistic data, the simulation environment needs to have an extremely high degree of realism and the ray tracing implementation needs to account for all the effects that occur in real \lidar sensors.
Additionally, every specific \lidar sensor has its own specifications and characteristics.
Of particular interest is the reflection behavior of the surfaces hit by the \lidar rays.
Materials with higher reflectivity result in a stronger signal, but can also cause light to be reflected away from the detector like a mirror.
This gives important cues, comparable to a grayscale camera image.
The respective measurement for this is the \emph{\lidar intensity}. 
While the strength of the detected signal also depends on the distance, sensors like the Velodyne HDL32E~\cite{velodyne_hdl32} return a distance adjusted value.

\begin{figure}[htp]
    \centering
\includegraphics[width=\textwidth, trim={2cm 0cm 6cm 0cm}]{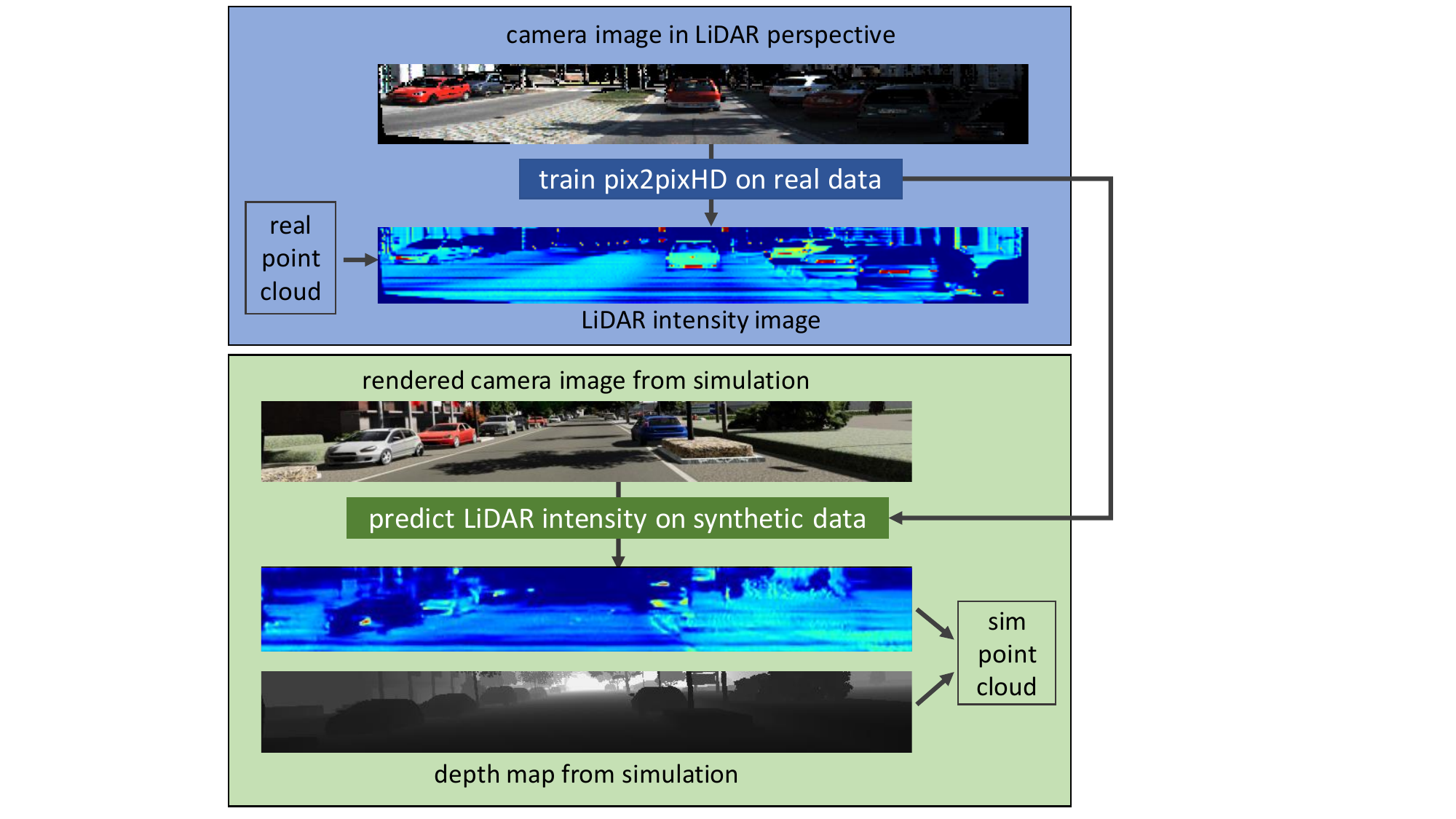}
\caption{Our \lidar point cloud simulation approach consists of two steps: First, the neural network is trained on real data (blue box). Second, it is fed with synthetic data to simulate point clouds (green box). We show \lidar intensity images using a color map for better visibility and keep depth maps in grayscale.}\label{fig:pipeline}
\end{figure}
An option to directly capture these effects is a learning-based approach that uses data from real test drives to simulate \lidar sensors.
If we can then simulate \lidar intensities, we can use them to filter a depth map, which is easily available in simulation environments, so that we can obtain a realistic synthetic \lidar point cloud in the end.
Figure~\ref{fig:pipeline} summarizes this approach. 
Distinctive parts in the point cloud then become visible like in real \lidar point clouds, which can improve performance in perception systems such as 3D object detection or semantic segmentation.
For the image translation task, we use pix2pixHD~\cite{wang2018pix2pixHD}, which follows the concept of General Adversarial Networks (GANs)~\cite{goodfellow2014generative}. 
We create a polar-grid image of the point cloud from the \lidar sensor.
To have a direct mapping, we also project the camera picture into the \lidar perspective so that each camera pixel corresponds to a \lidar point.
This process is described in greater detail in Section \ref{sec:architecture}.
As this results in two matching images, pix2pixHD can learn to translate from RGB to \lidar intensity.
While one use case for this would be to simulate \lidar point clouds from real test drives without the presence of an actual \lidar sensor, we focus on simulating realistic point clouds from synthetic data.
If we then use the simulated data as input for training real world systems, we obtain a \emph{Real2Sim2Real} pipeline that consists of the following steps:

\begin{itemize}
    \item \emph{Ground Truth:} Generate \lidar intensity images from the reflectivity measurements in \lidar point clouds (see Figure~\ref{fig:create_data_target}).
    \item \emph{Training Input:} Derive data from camera images (e.g., depth and segmentation maps) and project these images into perspective of the \lidar sensor (see Figure~\ref{fig:create_data_input}).
    \item \emph{Real2Sim:} Train a neural network on real data to predict \lidar intensity images. Using synthetic input data, the network can then simulate \lidar point clouds  (see Figure~\ref{fig:gen_cloud}).
    \item \emph{Sim2Real:}  Train real world perception systems (e.g., object detection) on \lidar point clouds from simulated data.
\end{itemize}

Our contribution in this is that we provide segmentation and depth data in addition to the camera data to pix2pixHD. 
By doing so, we can recover information about surfaces in the scene where no returns have been measured by the \lidar sensor, which enables performing the image translation task in the \lidar perspective.
Furthermore, we evaluate the quality of the simulated data beyond direct metrics between ground truth and generated data. Without this we would have to rely on visual comparisons as no \lidar ground truth point clouds are available for the synthetic data.
Thereby, we demonstrate that by leveraging real data, our simulation produces realistic \lidar point clouds.

After establishing the foundations in the related work, there are two core parts in our paper that are both evaluated, corresponding to our pipeline.
First, the simulation of \lidar intensities in Section~\ref{sec:simulation} and secondly, the generation of synthetic point clouds in Section~\ref{sec:detection}.
This is followed by a discussion of the implications for practical use cases in developing autonomous driving systems and ends with the conclusion.

\section{Related Work}
\paragraph{Simulation Environments} There are multiple capable simulation environments to test and develop advanced driving assistance and autonomous driving systems.
However, open source  tools like CARLA~\cite{dosovitskiy_carla_2017} or AirSim~\cite{shah_airsim_2017} only have basic sensor implementations, which are not sufficient for realistic simulation. 
Vista~\cite{amini2021vista} on the other hand employs data-based sensors, but is less focused on optimal integration into driving simulation workflows.
Commercial products, especially NVIDIA DRIVESim~\cite{nvidia_drivesim}, are  promising but usually employ high quality physics-based implementations.
Our approach on the other hand is supposed to be integrated easily without requirements to the simulation environment and should allow simple implementation of new sensors.
A further relevant approach is Pseudo-LiDAR~\cite{wang_pseudo-lidar_2018,you2020pseudo}. 
Here, predicted depth maps are converted and down sampled to mimic \lidar point clouds so that they can be used for object detection. 
We make use of this concept to generate point clouds from synthetic depth maps.
\paragraph{Image2Image Translation} For learning-based \lidar simulation, Image2Image translation has shown promising results.
This can be achieved effectively with GANs as shown by pix2pix~\cite{isola_image--image_2018}. 
In contrast to this, CycleGAN~\cite{zhu_unpaired_2020} can translate between two domains without the need for paired data.
Further improvements for paired images were made by pix2pixHD~\cite{wang2018pix2pixHD} and SPADE~\cite{park2019SPADE}.
Vid2Vid~\cite{wang_video--video_2018} improves the temporal consistency for a video sequences and World-Consistent Video2Video Translation~\cite{mallya2020world} further proposes solutions for long-term consistency.
Even though, test drive data is most of the time given as video sequences, approaches that process multiple consecutive frames simultaneously are problematic for \lidar intensity, because the reflected light strongly depends on the angle.
As this changes when the car is moving, warping between frames cannot be simply performed to improve the training.
Additionally, using the camera images as input already gives good data for temporal consistency.
\paragraph{\lidar Simulation} Methods for \lidar simulation based on machine learning use similar approaches. LiDARSim~\cite{manivasagam_lidarsim_2020} learns a mapping between a reconstructed dense point cloud and individual scans using a U-Net~\cite{ronneberger2015unet} based architecture.
However, this approach makes it difficult to integrate the \lidar simulation in simulation environments. 
There also is research regarding unpaired training, in this case using simulated unrealistic and real point clouds or point clouds from different sensors~\cite{sallab2019lidar}. 
While this offers great flexibility and various use cases, there is a conceptual problem to learn the \lidar behavior for specific materials or objects in the scene.

This behavior can be learned with paired GAN-based approaches that utilize camera images during training~\cite{Marcus_2022,guillard2022learning}. These approaches operate in the camera perspective, but the former further uses depth and segmentation images as additional input. The latter also simulates \lidar intensity in addition to dropping rays. 
Our approach incorporates both concepts but performs the image translation in the \lidar perspective, which has the distinct advantage that there is no need for upscaling the \lidar image to the camera resolution.
Aside from usual interpolation artifacts, \lidar data has the problem that gaps can either result from the sampling pattern of the sensor or the reflection properties of the surfaces hit.
Only blurring the \lidar points avoids larger interpolation errors, but also prevents the network to discern between the two different kinds of gaps. 
Fully matching the resolution of the camera, on the other hand, tends to also fill reflection gaps in the \lidar data.
As our method transforms the camera images into the \lidar perspective, it does not suffer from these problems.
In addition, training can be completed faster because only the lower resolution of the \lidar sensor is used.

\paragraph{Training Data}
For training the network, we focus on the KITTI~\cite{geiger_are_2012,geiger_vision_2013} dataset, as it includes various benchmarks and hand labeled data, which is helpful for training on high quality data, so the network can generalize well to the perfect synthetic samples and gives many options for evaluation. 
It uses a Velodyne~\cite{velodyne} HDL-64E \lidar sensor with 64 laser rows.
Furthermore, VKITTI~\cite{gaidon2016virtual,cabon2020vkitti2} creates a virtual model of five sequences of the KITTI dataset in a semiautomatic process. 
While the result is not a direct digital twin, as the geometry and object placement differs, it still provides a good starting point for comparative evaluations.
VKITTI provides stereo color images and corresponding depth, class segmentation and instance segmentation maps. It also contains the extrinsic and intrinsic camera parameters for each frame and bounding boxes for the cars.
Thus, we make this the basis for our synthetic data.

\section{Simulating \lidar Intensity}
\label{sec:simulation}
To leverage the simulation environment, a data-based sensor should make use of the available ground truth data.
The generation of realistic materials is often challenging and would be even more problematic to obtain from real world sequences.
Instead, we focus on readily available data: RGB renderings (camera images), depth, semantic and instance segmentation masks.
\subsection{Generating the Training Data}
\label{sec:architecture}
As these are supposed to be used during inference in the simulation, the training also needs to be performed with these modalities as input.
The ground truth for the training is obtained from only the \lidar data.

\paragraph{\lidar Data Processing}
\begin{figure}[htp]
    \centering
\includegraphics[width=\textwidth, trim={0cm 12cm 0 0cm}]{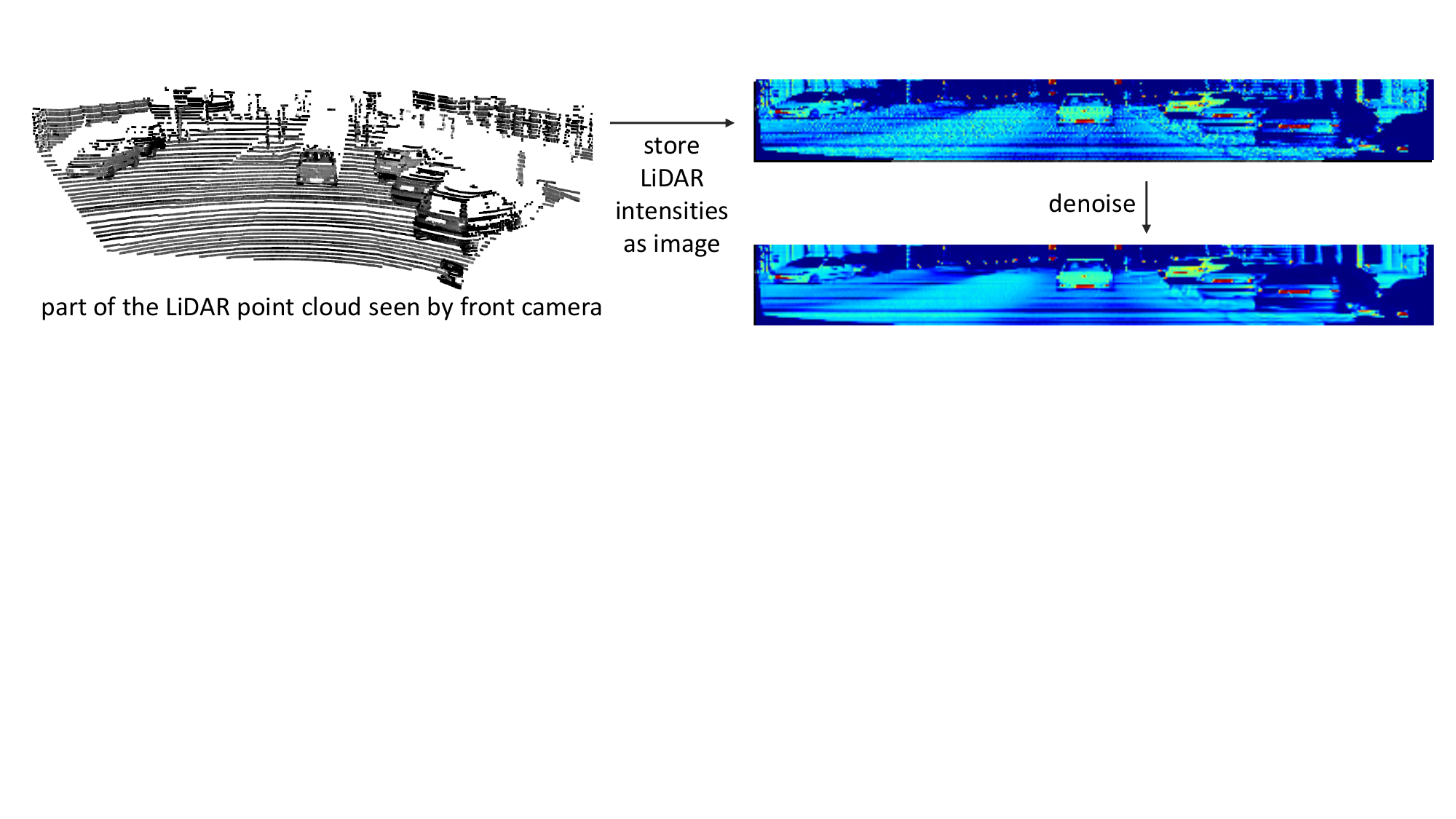}
\caption{Data processing and generation for \lidar intensity simulation with pix2pixHD.}\label{fig:create_data_target}
\end{figure}
To transform a \lidar point cloud into an image without sampling gaps, we must recover the way the points should have been measured.
Given an elevation angle and the azimuth (resulting from the laser arrays and rotating sensor), a \lidar point is located at the detected distance at these angle coordinates.
However, if we reverse this and try to map the individual points to pixel coordinates, we notice that this is not consistent. 
There can be gaps in the resulting image, which only should be the case when there is no detection at all. 
While the columns are quite accurate, the points are often registered at the wrong row.
To obtain a dense image, we exploit that the points in the KITTI files are ordered linewise.
Consequently, jumps between quadrants allow assignment of rows for each point. 

As our goal is to learn a mapping between camera and \lidar, only the overlapping parts can be used.
The camera only covers about 80 degree of the 360-degree \lidar panorama and the lowest rows of the sensor
are blocked by the ego car and also not in the camera perspective.
This effectively leads to a crop of 372 × 44 pixels, where each pixel gives a \lidar intensity and lies inside the camera view.
As a last step, we denoise the polar-grid \lidar images with LIDIA, a universal learned denoiser, as shown in Figure~\ref{fig:create_data_target}.
Even though noise is part of realistic \lidar intensity, we decided to exclude it from our experiments as good as possible to focus on how well the network can learn the ideal reflective properties of materials.
However, this could be analyzed in future work, as different materials could produce specific noise patterns in the measured intensities.


\paragraph{Camera Data Processing}
For the real input data, we need to start from the available sensors: \lidar and RGB camera.
\begin{figure}[htp]
    \centering
\includegraphics[width=\textwidth, trim={2cm 8cm 2,2cm 0cm}]{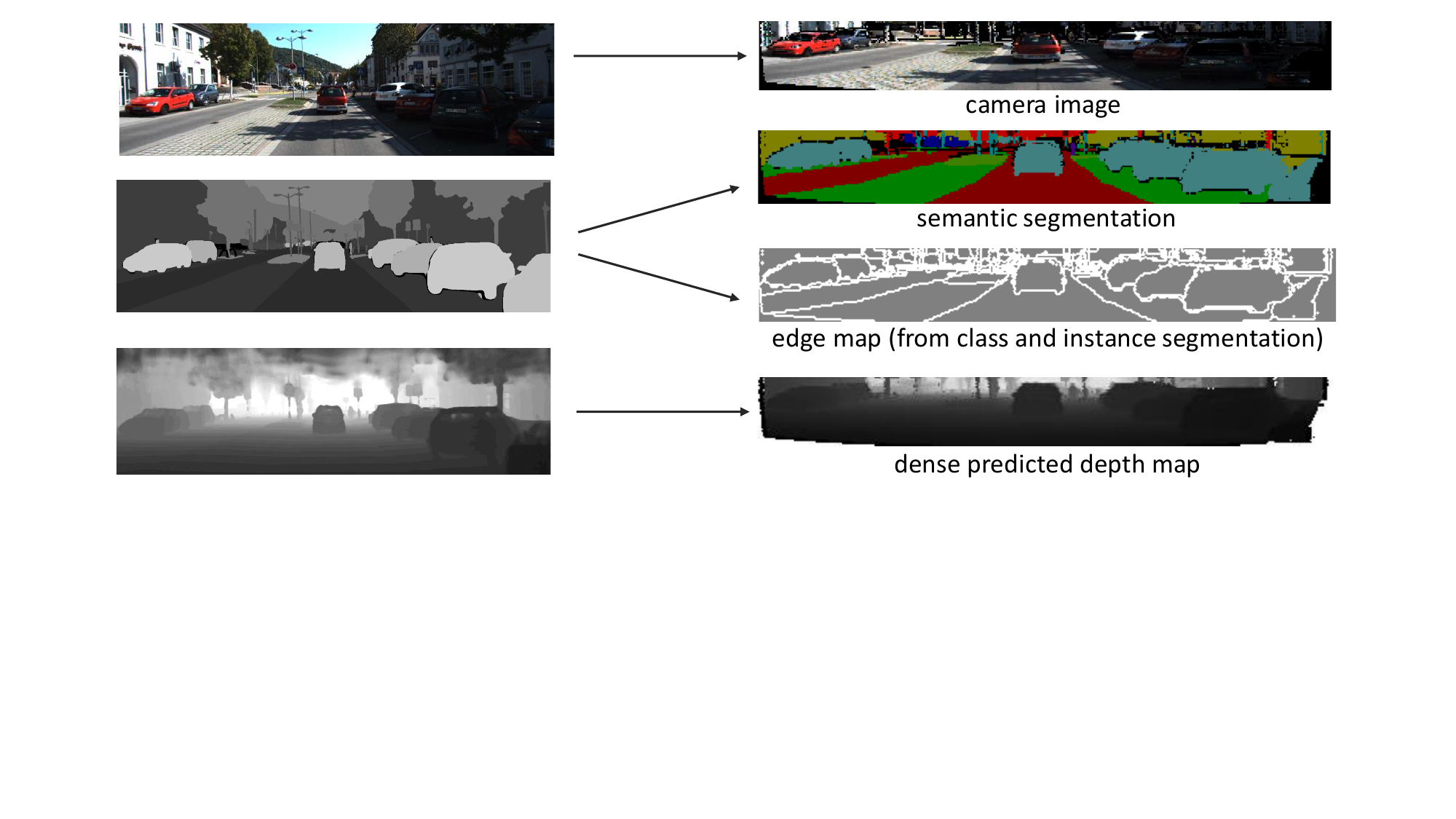}
\caption{Processing KITTI images for training: a projection from camera perspective to \lidar perspective is necessary.}\label{fig:create_data_input}
\end{figure}
Based on these, we can predict dense depth maps via the depth completion network PENet~\cite{hu2021penet}.
Depth completion means that the network uses the \lidar point cloud as basis to predict a dense depth map.
Hence, we not only use \lidar data in the ground truth but also as input.
This is no issue as we strictly want to predict reflection behavior in the form of \lidar intensity. 
Perfect depth maps are already available in the simulation environment for inference, later.
The predicted depth maps have a slightly lower resolution of 1216 × 352 pixels due to the training process, so we crop the original camera image as well.

For semantic and instance segmentation, we rely on manually labeled data, as given for the Semantic Instance Segmentation Evaluation Benchmark~\cite{Alhaija2018IJCV}. However, this could also be achieved with an appropriate neural network.
Instead of directly using the instance segmentation maps for training, we follow pix2pixHD and generate edge maps from the segmentation and instance data.

To project the camera data into the \lidar perspective, we also need depth information for every pixel.
Here, we can use the same predicted depth map that is one of the inputs for learning the intensity prediction.
Due to the offset between camera and \lidar sensor on the vehicle, there are parts in  the image that were not seen by the \lidar because of occlusions.
We detect these areas generously by checking whether multiple pixels from the camera image would fall into the same \lidar pixel coordinate and mask out occluded areas.
These masks are then added as \emph{don't care} labels to the segmentation masks.
When projecting the camera images, we also adjust the resolution to the \lidar crop~(372 × 44) so that each \lidar ray corresponds to one camera pixel, as shown in Figure~\ref{fig:create_data_input}.

\subsection{Results and Evaluation}
We use the pix2pixHD implementation from Imaginaire~\cite{Imaginaire} and train for 20 epochs with batch size 8, enable the local enhancer network, but deactivate horizontal image flipping during training.
Otherwise, we keep the default configuration.

\begin{figure}[htp]
    \centering
    \begin{subfigure}[b]{0.49\textwidth}
    \centering
    \includegraphics[width=\textwidth, trim={4,2cm 0cm 11cm 0cm}, clip=true]{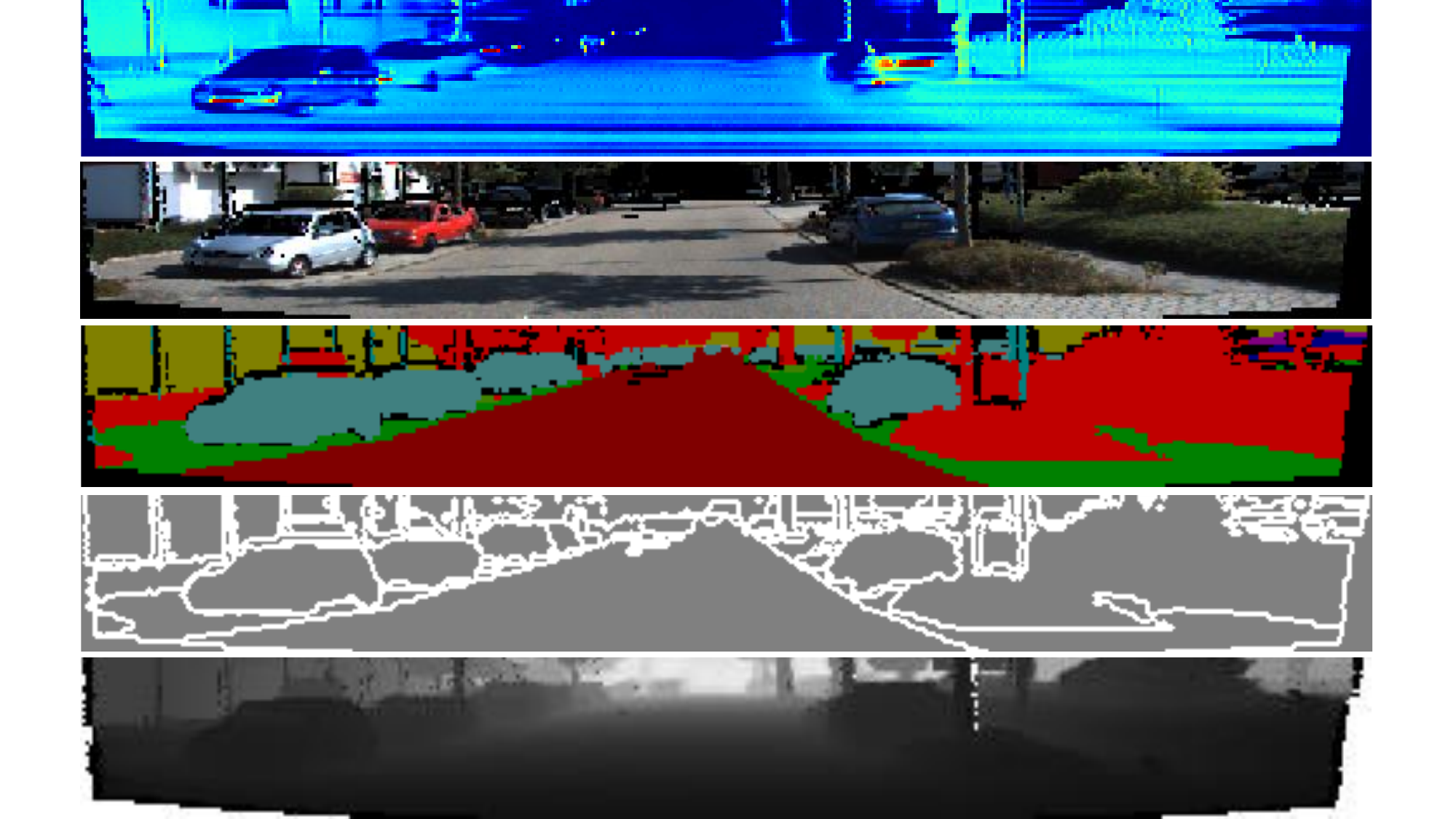}
        \caption{First row: ground truth \lidar intensity, below: input images. Images are cropped slightly for better visibility.}\label{fig:eval_real_gt}
    \end{subfigure}
       \hfill
 \begin{subfigure}[b]{0.49\textwidth}
   \centering
   \includegraphics[width=\textwidth, trim={4,2cm 0cm 11cm 0cm}, clip=true]{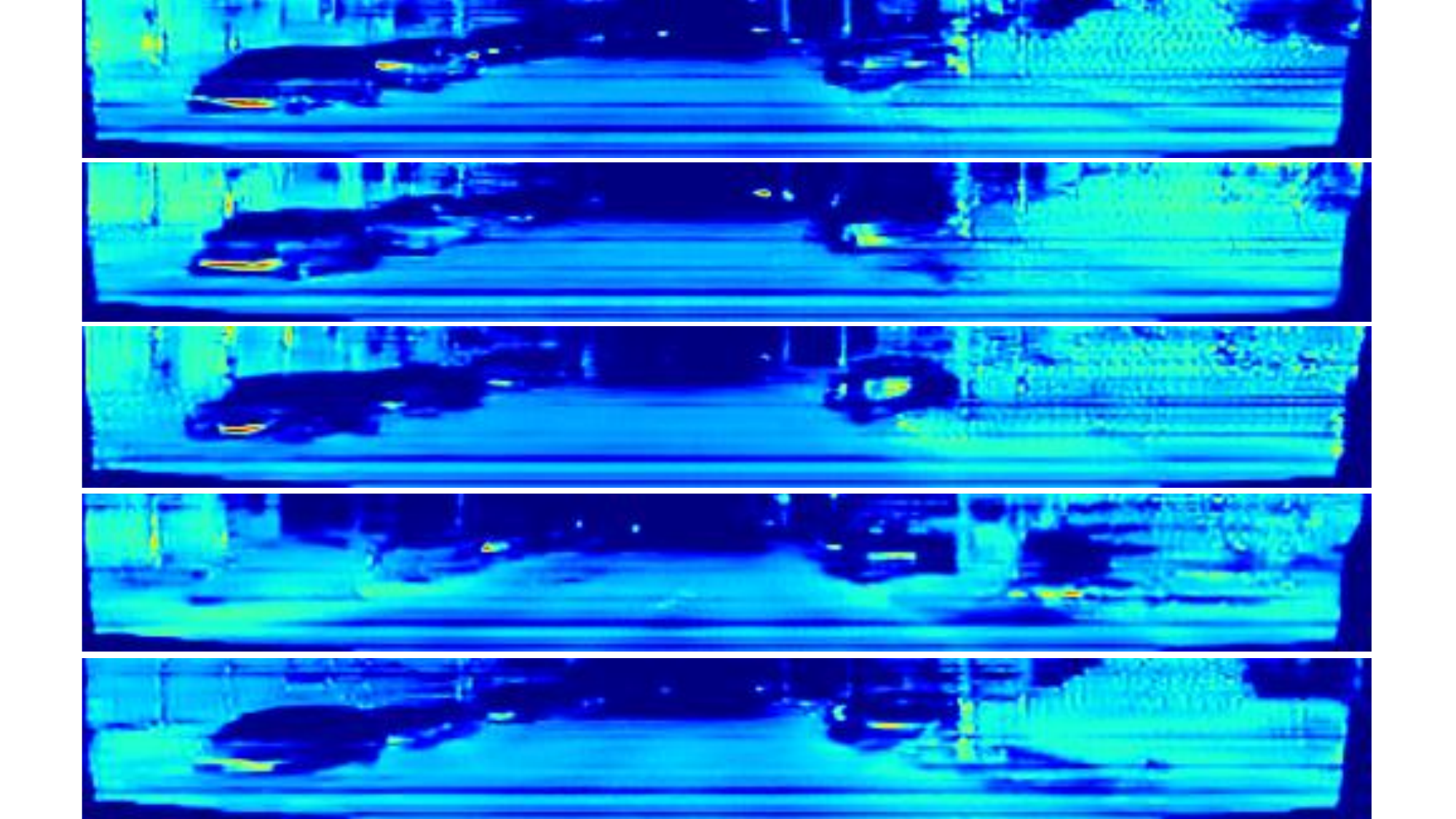}
    \caption{First row: \lidar intensity prediction from combined input, below: predictions from individually trained networks.}\label{fig:eval_real_out}
    \end{subfigure}
\caption{Qualitative evaluation with sequence from the KITTI validation split using different input data configurations.}\label{fig:eval_real}

\end{figure}

We train on 16 sequences (about 6000 images) of the training and validation data of the segmentation benchmark and use the five sequences (about 2000 images) that are also modeled in VKITTI for evaluation.  
In particular, we compare between different runs of camera images and derived images individually, as well as one variant where we use all available input data.
Images are always scaled to a resolution of 512 × 64. 
As metric, we use the Fréchet inception distance (FID)~\cite{heusel2018gans}, which has proven to be an effective way to compare similarity between generated and real images.
The numerical results can be seen in Table~\ref{tab:fid} and the corresponding images in Figure \ref{fig:eval_real}. 
    \begin{table}
      \centering
        \caption{FID between ground truth and predictions on validation set at epoch 20.}\label{tab:fid}
        \begin{tabular}{l|c|c|c|c|c}

        Input Data    & RGB   & Semantic  & Edges  & Depth & Combined \\

        \hline
        FID     & 89.44    & 77.01        & 102.61     & 81.43    & 73.81\\
  
        \end{tabular}
    \end{table}
    
In general, the numbers show a clear picture of what the network learns from the data and how this generalizes to unseen data. 
Class segmentation and depth provide information that applies very similarly to materials and geometry across scenes and thus promote generalization capabilities. 
The same is true for the edge maps in theory, albeit they contain considerably less information by themselves.
The RGB camera images behave very differently.
They are full of high frequent information, which also makes it very hard for the network to apply on new data.
The value of including RGB images is not diminished by this, as they provide valuable information for simulating intensities that correspond to the specific materials in the frame and also help with temporal stability.

\paragraph{Generalization on Synthetic Data}
\begin{figure}[htp]
    \centering
\includegraphics[width=\textwidth, trim={2,2cm 8cm 2,2cm 0cm}]{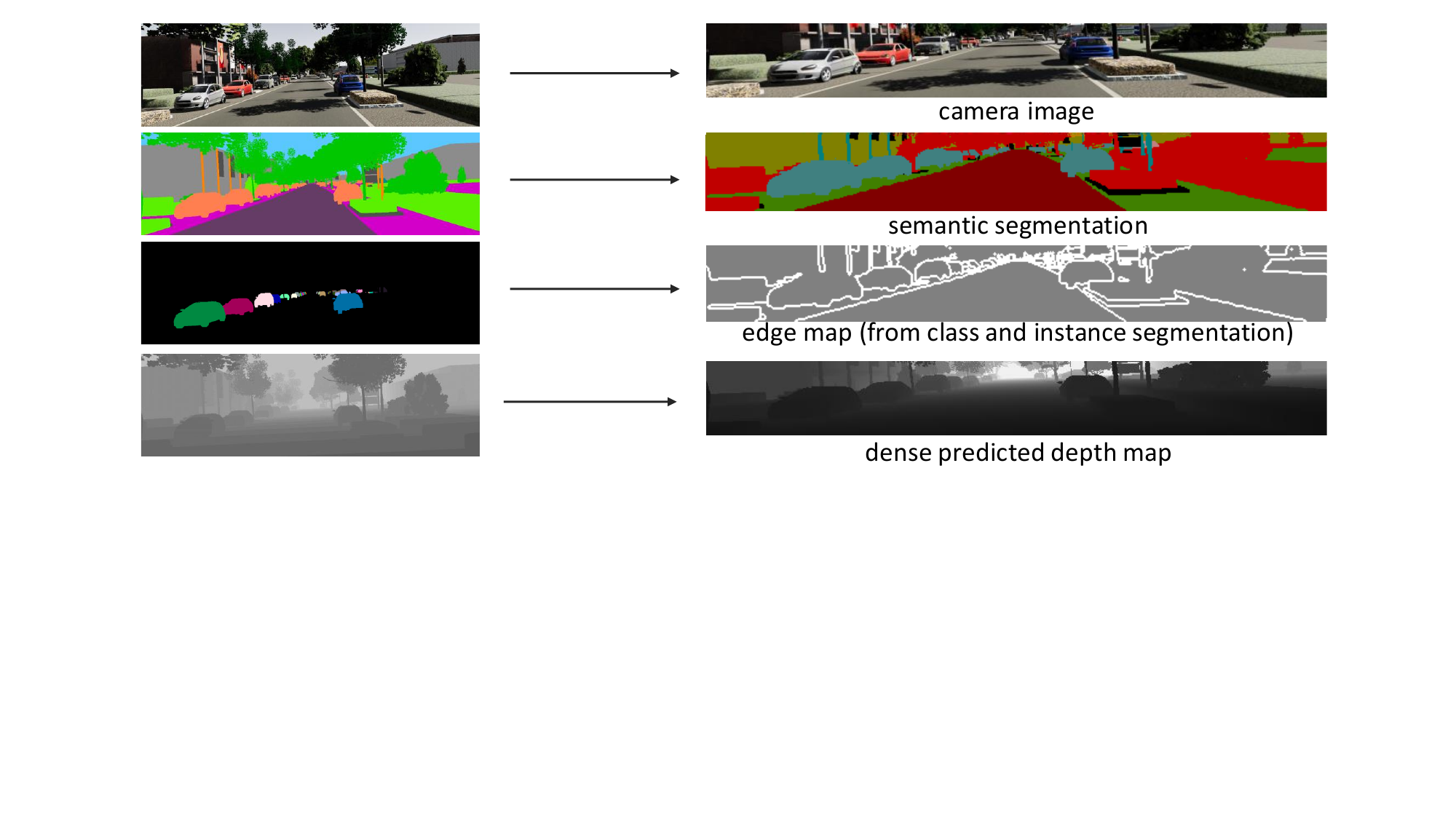}
\caption{Processing VKITTI images for training: instead of projection, we directly downsample the images to a resolution of 372 × 44 pixels.}\label{fig:create_syn}
\end{figure}

VKITTI provides the necessary data, but it has to be adjusted slightly to adhere to the training format.
The depth maps need to be scaled into the same range as in the completed depth images.
The color-based classes of VKITTI need to be mapped to KITTI class identifiers.
The color-based instance segmentation maps of VKITTI need to be mapped to subsequent numbers.
The conversion results are shown in Figure~\ref{fig:create_syn}.
Even though the training is performed on data converted into the \lidar perspective, the network itself  learns a mapping between camera-derived data and \lidar intensity. This means that we can directly evaluate the network on the synthetic images. As long as the aspect ratio and resolution are comparable, the generated data is plausible.
Without further projection, this would simulate a \lidar sensor that has the same position as the camera and a resolution corresponding to the training resolution.
A higher resolution \lidar could be simulated by simply sampling a different set of pixels from the source image.

\begin{figure}[htp]
    \centering
    \begin{subfigure}[b]{0.49\textwidth}
    \centering
    \includegraphics[width=\textwidth, trim={4,2cm 0cm 11cm 0cm}, clip=true]{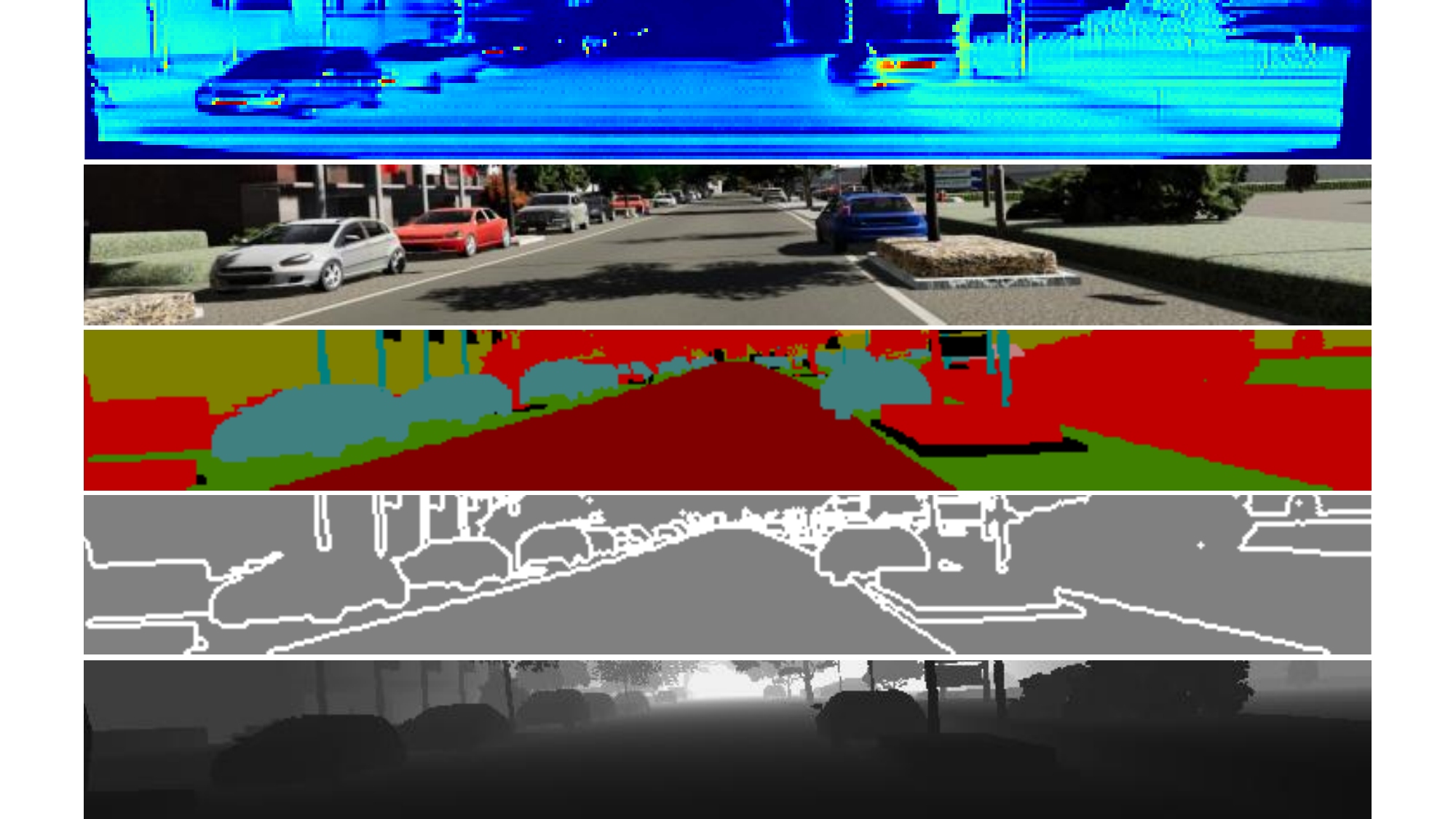}
        \caption{First row: ground truth \lidar intensity (KITTI), below: input images. Images are cropped slightly for better visibility.}\label{fig:eval_syn_gt}
    \end{subfigure}
       \hfill
 \begin{subfigure}[b]{0.49\textwidth}
   \centering
   \includegraphics[width=\textwidth, trim={4,2cm 0cm 11cm 0cm}, clip=true]{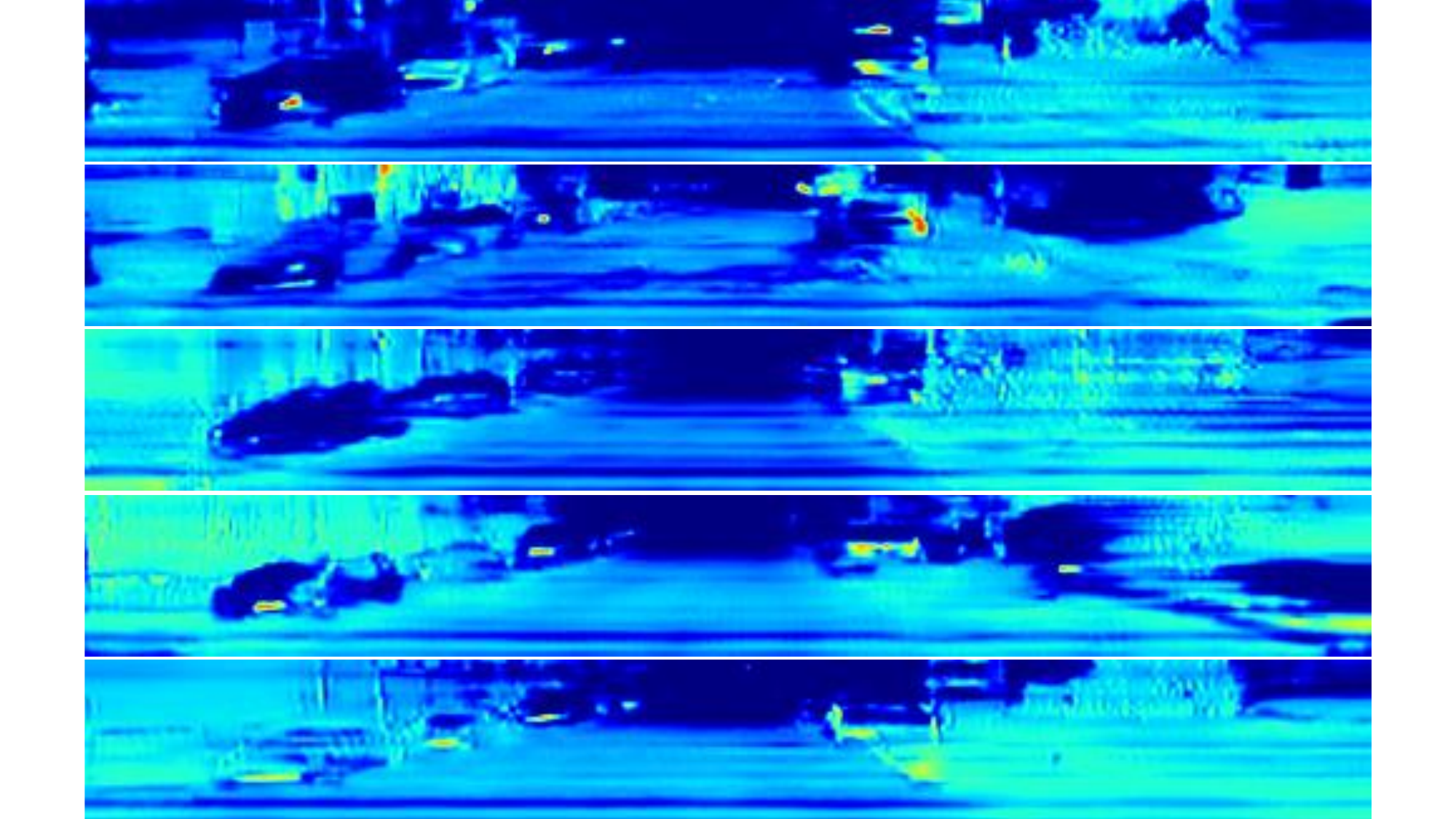}
    \caption{First row: \lidar intensity prediction from combined input, below: predictions from individually trained networks.}\label{fig:eval_syn_out}
    \end{subfigure}
\caption{Qualitative evaluation with the VKITTI sequence corresponding to the KITTI sequence in Figure \ref{fig:eval_real}
using different input data configurations.}\label{fig:eval_syn}
\end{figure}

We show the results of the networks trained on the different kinds of real world input images with synthetic data input from VKITTI in Figure~\ref{fig:eval_syn}, using the corresponding KITTI validation sequence.
This is the same sequence for which we have already showed the predicted intensities in Figure~\ref{fig:eval_real}, but there are notable differences between the real and the synthetic scene itself.
Still, the predictions align reasonably well, especially considering that there is a domain gap, not only between camera and rendered image but also between perfect and predicted depth maps as well as between the segmentation maps because of different class labels. 
However, we cannot evaluate the performance quantitatively in the same way.
We provide a solution to this by creating point clouds from the predicted intensity and analyze them for the task of training object detection.

\section{Object Detection with Simulated \lidar Point Clouds}\label{sec:detection}
To further evaluate the realism of the simulated point clouds, we train an object detection network on VKITTI and evaluate the performance on real KITTI data. 
Furthermore, we also evaluate a model trained on real KITTI data on the synthetic point clouds.
In each case, the network architecture is Voxel-R-CNN~\cite{deng2020voxel}. 
The unstructured point clouds are first encoded into voxels. This means that it should be rather robust against sensor noise or aliasing from the depth projection in synthetic point clouds.

\paragraph{Generating Point Clouds from Synthetic Data}

 In order to use the VKITTI dataset analogously to the KITTI dataset, we have to generate the point clouds from the depth maps and reconstruct the calibration files. 
For the image files, we use the RGB images of the left VKITTI camera (Camera\_0) and generate the 3D and 2d bounding box information in the correct format with the information from  the supplied text files.
We remove bounding boxes that exceed a maximum distance of 80 meters and exclude scenes that contain no objects completely.

In contrast to the real KITTI data, we define no objects as completely visible, because VKITTI gives very accurate occlusion values, which however do not exceed a visibility of 90 percent.
This has no effect on the training process itself, only during evaluating on the VKITTI sequences there will be no easy category for evaluating object detections.
Instead, these samples will be part of the moderate difficulty.

\begin{figure}[htp]
    \centering
\includegraphics[width=\textwidth, trim={2,5cm 11cm 2,8cm 0cm}]{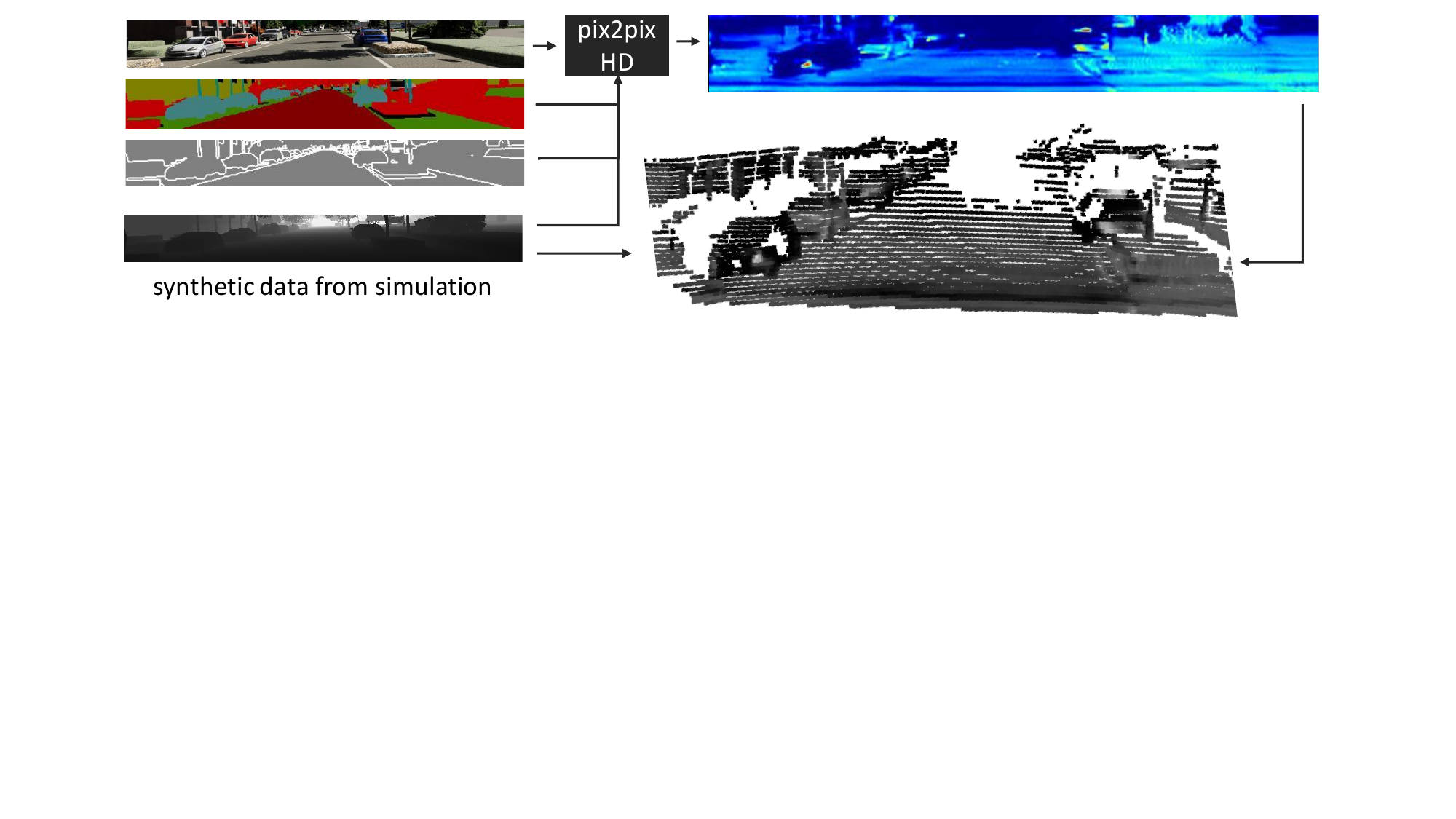}
\caption{Generating synthetic point clouds from depth maps and predicted \lidar intensity.}\label{fig:gen_cloud}
\end{figure}

For a specific \lidar sensor position, there are three possibilities.
 First, the point cloud generated from the depth map and the predicted intensity can be projected into the \lidar coordinate system.
Second, equivalent to the training procedure, the camera data can be projected first and in this form be fed into the network.
Last, an additional camera could be simulated at the desired \lidar position.
This should in most cases be the most accurate solution, but cannot be applied on already existing data like VKITTI.
For our evaluation, we choose the first option and convert the depth maps to point clouds following Pseudo-LiDAR~\cite{wang_pseudo-lidar_2018}. We make one important modification: the intensity of \lidar points is set to 0 instead of 1. Training on real point clouds and evaluating on synthetic point clouds with intensity 1 causes the detection performance to collapse because the detector then considers every point as retro-reflective.
Adding the intensity simulation enables us to assign an intensity to points or drop them completely with low intensity.
We apply this conservatively and only drop points with 0 intensity (see Figure~\ref{fig:gen_cloud}).
We only use the network once for each frame and upscale the low resolution intensity map to the cropped KITTI resolution of  1216 × 352, which assigns an intensity of 0 to the border pixels from the original KITTI and VKITTI resolution of 1242 × 375 pixels.
This allows using the full available resolution of the depth map.
After generating a point cloud from it, we sparsify the point cloud according to the desired sensor specifics following the Pseudo-LiDAR method~\cite{you2020pseudo}.

Generating point clouds from VKITTI results in about 2000 samples for training. We generate three different sets of simulated VKITTI point clouds and compare them against real KITTI point clouds, following the procedure above: one with 32 lines and two with 64 lines, once without further processing and once with dropping points according to the predicted intensity.

\begin{figure}[htp]
    \centering
    \begin{subfigure}[b]{.49\textwidth}
    \centering
    \includegraphics[width=\textwidth, trim={3cm 0cm 4cm 0cm}, clip=true]{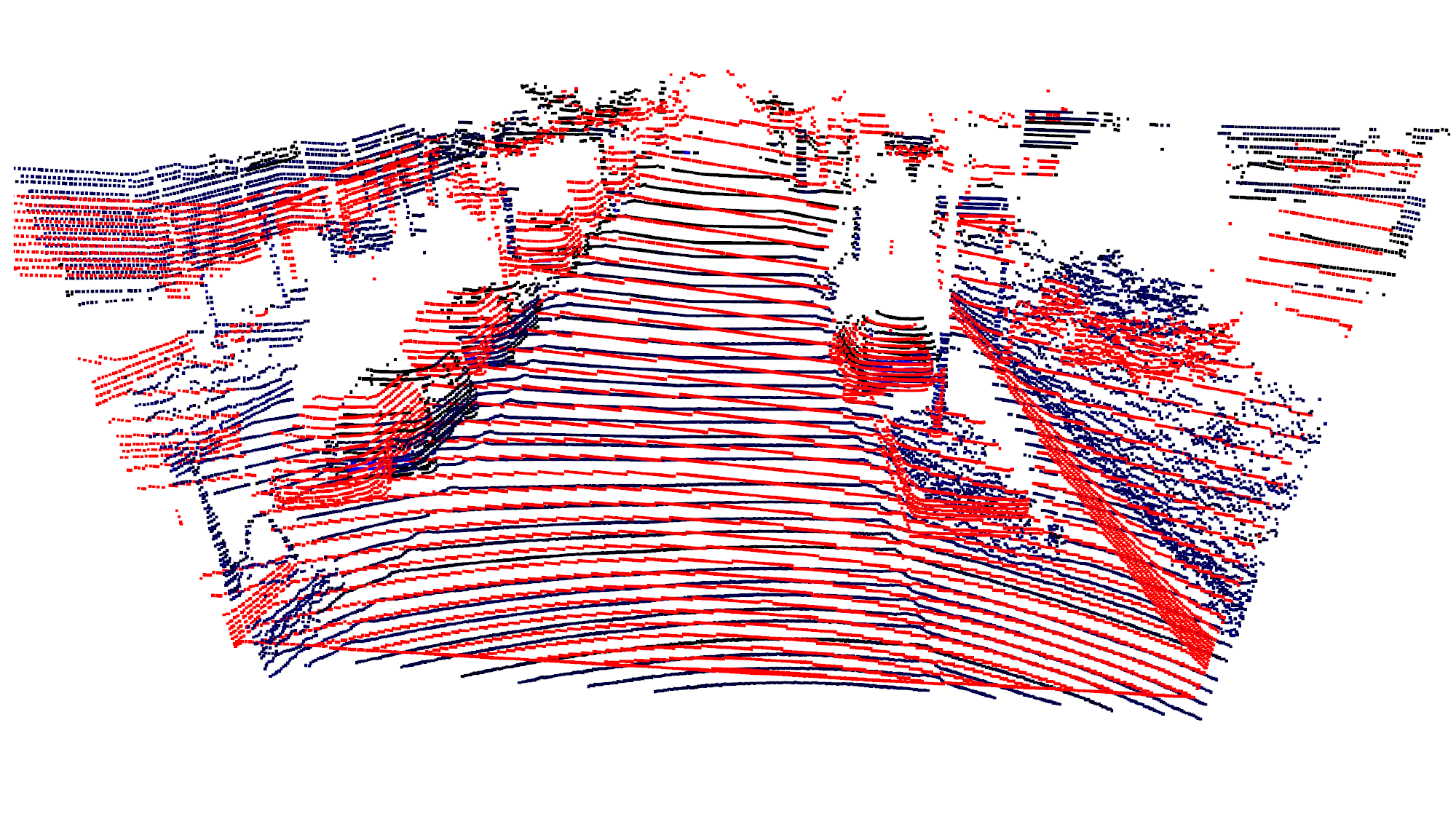}
        \caption{Real point cloud (blue) and point cloud from projected synthetic depth (red).}
    \end{subfigure}
       \hfill
 \begin{subfigure}[b]{.49\textwidth}
   \centering
   \includegraphics[width=\textwidth, trim={4cm 0cm 4cm 0cm}, clip=true]{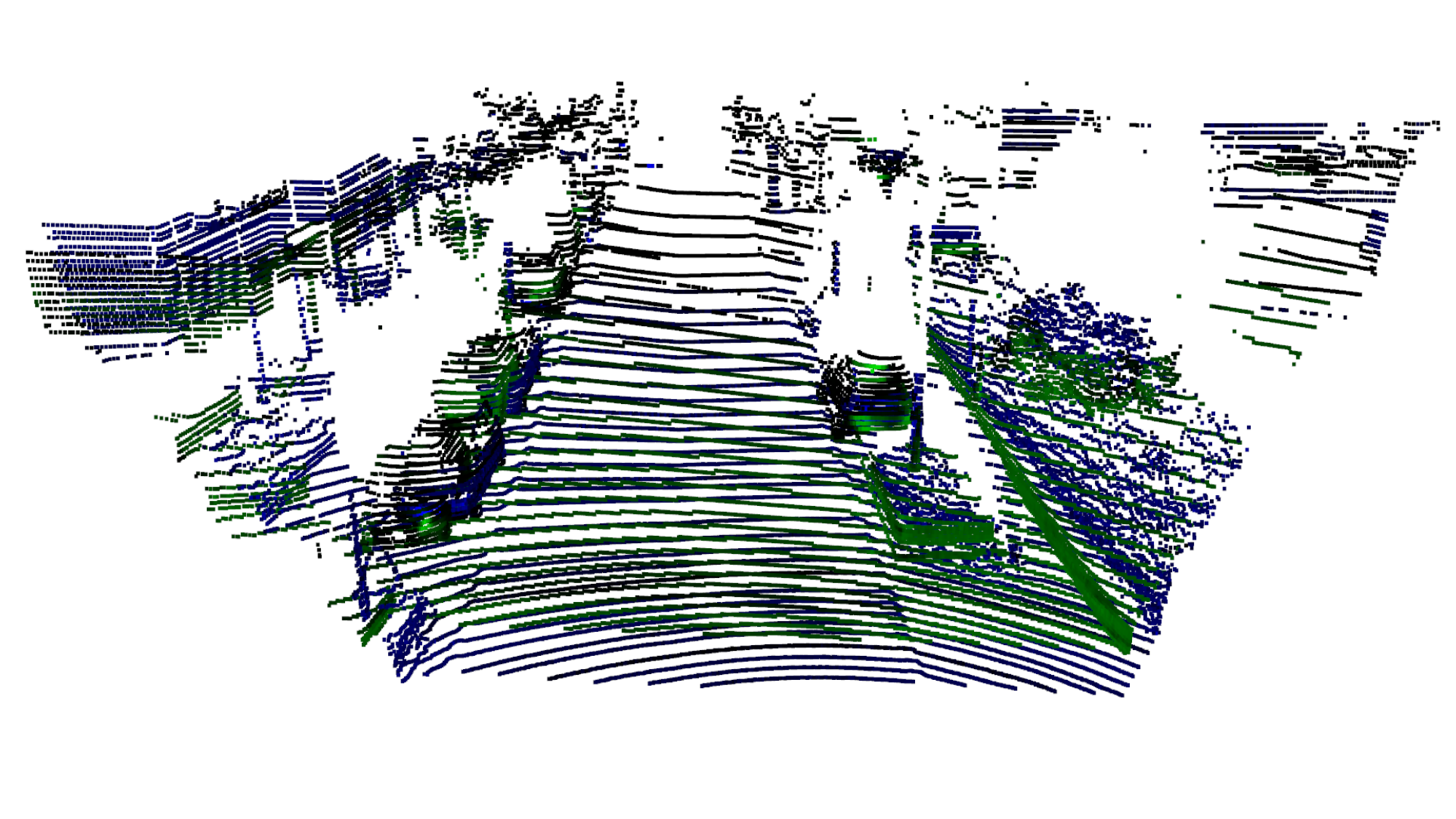}
    \caption{Using simulated intensity in addition to the depth (green).}\label{fig:sim_cloud}
    \end{subfigure}
\caption{Comparison between synthetic and real point clouds. Using the same image area as during training causes the lowest \lidar  rays to be cut off.}\label{fig:compare_clouds}
\end{figure}

The reason for also evaluating a variant with only 32 lines is to gain insights whether the observed effects could also simply result from having less points overall.
Visualizations of the 64 line point clouds are given in Figure~\ref{fig:compare_clouds}. This also shows the geometry discrepancy between VKITTI and KITTI. However,
this is not a problem for comparing the general object detection performance, because we do not compare point clouds individually.
The point cloud with simulated intensity on the right (Figure~\ref{fig:sim_cloud}) matches the real point cloud structure, having more sparsity in the distant parts and around reflective and transparent surfaces.

To analyze the results, we follow the KITTI object detection benchmark, which comes with 7481 point cloud and labels that are roughly split in half for training and evaluation.
It divides the samples for evaluation into easy, moderate and hard depending on the size, occlusion and truncation in the camera image.
The network outputs 3D and 2D bounding boxes, a top-down bird's eye view detection (bev) and average orientation similarity (aos), while we follow the KITTI object evaluation to calculate the detection scores.
We always train Voxel-RCNN for 80 epochs for stable results and average the last 5 epochs to account for remaining fluctuations between epochs.


\paragraph{Evaluation on KITTI}

\begin{table}[h!t]
    \centering
    \begin{tabular}{lcccccc}\toprule
    easy & 3D@0.70 & bev@0.70 & 3D@0.50 & bev@0.50 & 2D & aos\\\midrule
    VKITTI (64 Lines) & 63.78 & 75.32 & 79.93 & 80.24 & 79.24 & 78.58 \\
    VKITTI (32 Lines) & 61.4 & 73.61 & 78.28 & 78.62 & 77.3 & 76.62 \\
    IntensitySim (64) & 63.22 & 80.33 & 85.69 & 86.15 & 83.83 & 82.1 \\
    KITTI & 91.94 & 95.36 & 98.51 & 98.55 & 98.46 & 98.43 \\
    \end{tabular}
    
    \begin{tabular}{lcccccc}\toprule
    moderate & 3D@0.70 & bev@0.70 & 3D@0.50 & bev@0.50 & 2D & aos\\\midrule
    VKITTI (64 Lines) & 50.5 & 62.43 & 68.54 & 68.91 & 66.59 & 65.05 \\
    VKITTI (32 Lines) & 47.56 & 59.57 & 65.16 & 65.51 & 63.73 & 61.81 \\
    IntensitySim (64)  & 51.74 & 70.77 & 75.51 & 79.71 & 72.78 & 70.54 \\
    KITTI & 82.9 & 91.19 & 94.78 & 95.46 & 94.65 & 94.52 \\
    \end{tabular}

    \begin{tabular}{lcccccc}\toprule
    hard & 3D@0.70 & bev@0.70 & 3D@0.50 & bev@0.50 & 2D & aos\\\midrule
    VKITTI (64 Lines) & 48.7 & 62.07 & 69.01 & 70.02 & 66.84 & 65.05 \\
    VKITTI (32 Lines) & 45.2 & 58.15 & 64.04 & 65.42 & 62.9 & 60.93 \\
    IntensitySim (64) & 50.27 & 69.74 & 74.76 & 79.07 & 71.95 & 69.42 \\
    KITTI & 80.45 & 88.94 & 94.49 & 94.59 & 92.34 & 92.14 \\
    \end{tabular}
    \caption{Voxel R-CNN Objected detection evaluation on KITTI point clouds, the first column specifies the data set used for training.}
    \label{tab:kitti_eval}
\end{table}

In the first experiment, we have trained a Voxel R-CNN on KITTI point clouds as well as all types of our synthetic point clouds. All of these models are then evaluated on the real data, see Table~\ref{tab:kitti_eval}.
First, we notice that there is a significant gap between the models trained on real and synthetic data for generalization.
The difficult part is to determine how much can be attributed to the individual scene differences, point cloud structure or data set selection.
We see that simply reducing the point density to 32 lines decreases performance. In theory, an increase could have been possible if there were no sparse car samples in the synthetic for the network to learn to detect them in real data.
The \emph{IntensitySim} point clouds, on the other hand, also consist of fewer points, but the detection performance increases for almost every metric, which means that points have been dropped correctly.
Only for the easy samples, the performance slightly decreases compared to the regular VKITTI point cloud. 
This is not unexpected because the detection of these rely on a representative number of dense car samples in the training set.

\paragraph{Training on KITTI and Evaluation on VKITTI}

\begin{table}[h!t]
    \centering
    \begin{tabular}{lcccccc}\toprule
    moderate & 3D@0.70 & bev@0.70 & 3D@0.50 & bev@0.50 & 2D & aos\\\midrule
    VKITTI (64 Lines) & 88.78 & 95.79 & 98.94 & 99.25 & 98.59 & 97.89 \\
    VKITTI (32 Lines) & 72.93 & 90.74 & 94.3 & 97.54 & 89.78 & 88.48 \\
    IntensitySim (64)  & 86.33 & 99.11 & 99.89 & 99.92 & 99.78 & 98.35 \\
    KITTI & 82.9 & 91.19 & 94.78 & 95.46 & 94.65 & 94.52 \\
    \end{tabular}
    
    \begin{tabular}{lcccccc}\toprule
    hard & 3D@0.70 & bev@0.70 & 3D@0.50 & bev@0.50 & 2D & aos\\\midrule
    VKITTI (64 Lines) & 77.18 & 87.84 & 92.47 & 93.94 & 90.59 & 89.51 \\
    VKITTI (32 Lines) & 61.2 & 80.42 & 85.12 & 89.96 & 79.68 & 77.85 \\
    IntensitySim (64)  & 81.19 & 95.02 & 96.89 & 96.97 & 95.97 & 94.37 \\
    KITTI & 80.45 & 88.94 & 94.49 & 94.59 & 92.34 & 92.14 \\
    \end{tabular}
    \caption{Voxel R-CNN Objected detection evaluation on VKITTI point clouds, the first column specifies the data set used for training (no samples for easy difficulty).}
    \label{tab:kitti_train}
\end{table}

The second experiment evaluates the detection performance of a model trained on real KITTI data when applied on synthetic data, in our case the VKITTI point clouds.
With the perfect synthetic data, object detection is often too exact in simulation environments.
On the other hand, there might be a domain gap, that prevents the network from reaching the optimal performance.
The data in Table~\ref{tab:kitti_train} shows that the latter largely is the case.
This is significant as the performance increases despite having less points per sample for detection.
Furthermore, it indicates that the intensity was simulated realistically.
Again, simply reducing the point cloud resolution causes the network trained on higher resolution KITTI point clouds to miss many objects.

\section{Discussion}
When designing a system that in the end is supposed to bring autonomous driving functions to the street, we carefully have to consider the limitations of the simulation capabilities.
In our case, we want the network to learn a mapping between data from camera and \lidar.
Yet, the information given in single camera frames cannot determine the reflection behavior of the materials seen by the \lidar. Materials exist, that look exactly the same in the camera images but not for the \lidar.
Trying to test specific situations with such a sensor in the simulation is therefore a problematic approach when we want to replicate real world behavior exactly.
Conversely, the data-based sensor can be used to generate diverse virtual training data.

Going back from this general observation to the data of our experiments, we consider improving the training data as very important.
For the real data set, higher resolution \lidar sensors could improve the training drastically and the data projection methods can be optimized.
Here, both the predicted depth and the occlusion masking can introduce artifacts in the training data.
While VKITTI offered interesting insights because of its similarity, having a dynamic simulation environment would allow greater flexibility for generating synthetic training data and thus also experiments that investigate how to optimize generalization from synthetic to real data, in particular, when it is based on real data.

\section{Conclusion}
We have proposed a pipeline to simulate \lidar point clouds, including intensity, from real data and validate the realism by observing the generalization of an object detection network. In particular, we employed pix2pixHD for Image2Image translation and Voxel-R-CNN for object detection using KITTI point clouds and synthetic data from VKITTI as starting points. Converting these appropriately resulted in reasonable visual quality and FID scores, showing that using all available data as input performed best. The data from the object detection experiments strongly indicates that our \lidar simulation approach can increase the realism of the synthetic point clouds, thus creating a valuable starting point for evaluating different data configurations.

%
%
%
\bibliographystyle{splncs04}
\bibliography{main}
%

\end{document}